\documentclass[10pt,twocolumn,letterpaper]{article}

\usepackage[pagenumbers]{cvpr} 

\usepackage{graphicx}
\usepackage{amsmath}
\usepackage{amssymb}
\usepackage{booktabs}

\usepackage[pagebackref,breaklinks,colorlinks]{hyperref}

\usepackage[capitalize]{cleveref}
\crefname{section}{Sec.}{Secs.}
\Crefname{section}{Section}{Sections}
\Crefname{table}{Table}{Tables}
\crefname{table}{Tab.}{Tabs.}

\newcommand{\figref}[1]{Fig.~\ref{#1}}
\newcommand{\tabref}[1]{Table~\ref{#1}}

\def\eg{\emph{e.g}\onedot} 
\def\ie{\emph{i.e}\onedot} 
 
\def\etc{\emph{etc}\onedot}

\makeatother

\newcommand{\bftab}{\fontseries{b}\selectfont}

\def\cvprPaperID{1190} 
\def\confName{CVPR}
\def\confYear{2023}

\begin{document}
\newcommand{\ch}[1]{\textcolor{orange}{[Chris: #1]}}

\title{FSID: Fully Synthetic Image Denoising via Procedural Scene Generation}

\author{Gyeongmin Choe, Beibei Du, Seonghyeon Nam, Xiaoyu Xiang, Bo Zhu, Rakesh Ranjan \vspace{3mm}\\Meta Reality Labs}

\maketitle

\begin{abstract}
For low-level computer vision and image processing ML tasks, training on large datasets is critical for generalization. However, the standard practice of relying on real-world images primarily from the Internet comes with image quality, scalability, and privacy issues, especially in commercial contexts. To address this, we have developed a procedural synthetic data generation pipeline and dataset tailored to low-level vision tasks.
Our Unreal engine-based synthetic data pipeline populates large scenes algorithmically with a combination of random 3D objects, materials, and geometric transformations. Then, we calibrate the camera noise profiles to synthesize the noisy images. From this pipeline, we generated a fully synthetic image denoising dataset (FSID) which consists of 175,000 noisy/clean image pairs. 
We then trained and validated a CNN-based denoising model, and demonstrated that the model trained on this synthetic data alone can achieve competitive denoising results when evaluated on real-world noisy images captured with smartphone cameras.
\end{abstract}
\vspace{-3mm}

\section{Introduction}
\label{sec:intro}
With the recent growth of data-driven approaches for computer vision and image processing, having an accurate and large amount of training dataset~\cite{ILSVRC15,openimages,lin2014microsoft} for training and validation has become more critical. Especially, low-level computer vision applications such as image denoising~\cite{chen2022simple,young2022feature}, super resolution~\cite{zhang2018image,xiang2020zooming,liang2021swinir,chan2022basicvsr++}, video deblurring~\cite{liang2022vrt,yang2022motion,liang2022rvrt} \etc require sufficient texture or color variations in patches and pixels in an image for training a model. For example, image denoising requires accurately aligned pair images of clean ground truth (GT) frame and noisy frame to train the learning-based denoising model. Capturing the real noisy/clean image pairs, in reality, is challenging. Because firstly, capturing real images takes much time and effort and is limited to human labor. Although a representative conventional denoising dataset~\cite{Abdelhamed_2018_CVPR} presents high-quality real-captured noisy images, the scene variation is limited to the tens of scenes, also the total number of frames is limited. To overcome this limitation, there has been some research on doing noise profile calibration and injecting the calibrated noise profile onto the captured real images~\cite{wang2020}. This method models the noise profile of some smartphone cameras, and injects the noise profile onto a high-exposure image. This high-exposure image is regarded as a clean ground truth image. Although this method demonstrates the synthetically generated noise images practically denoise the noisy images, still it requires real smartphone camera-captured images in high exposure setting to obtain the clean ground truth images where the calibrated noise profile should be injected on. Also, the alignment of the noisy/clean images is carefully considered, and obtaining an accurate ground truth is not easy (Note that even high exposure images might contain inherent noise.) In our paper, our method takes a further step: we present a fully synthetic denoising pipeline, which solely uses synthetically generated clean ground truth images as well as synthetically generated noisy images to train the denoising neural networks.

\begin{figure}[t]
    \includegraphics[width=0.48\textwidth]{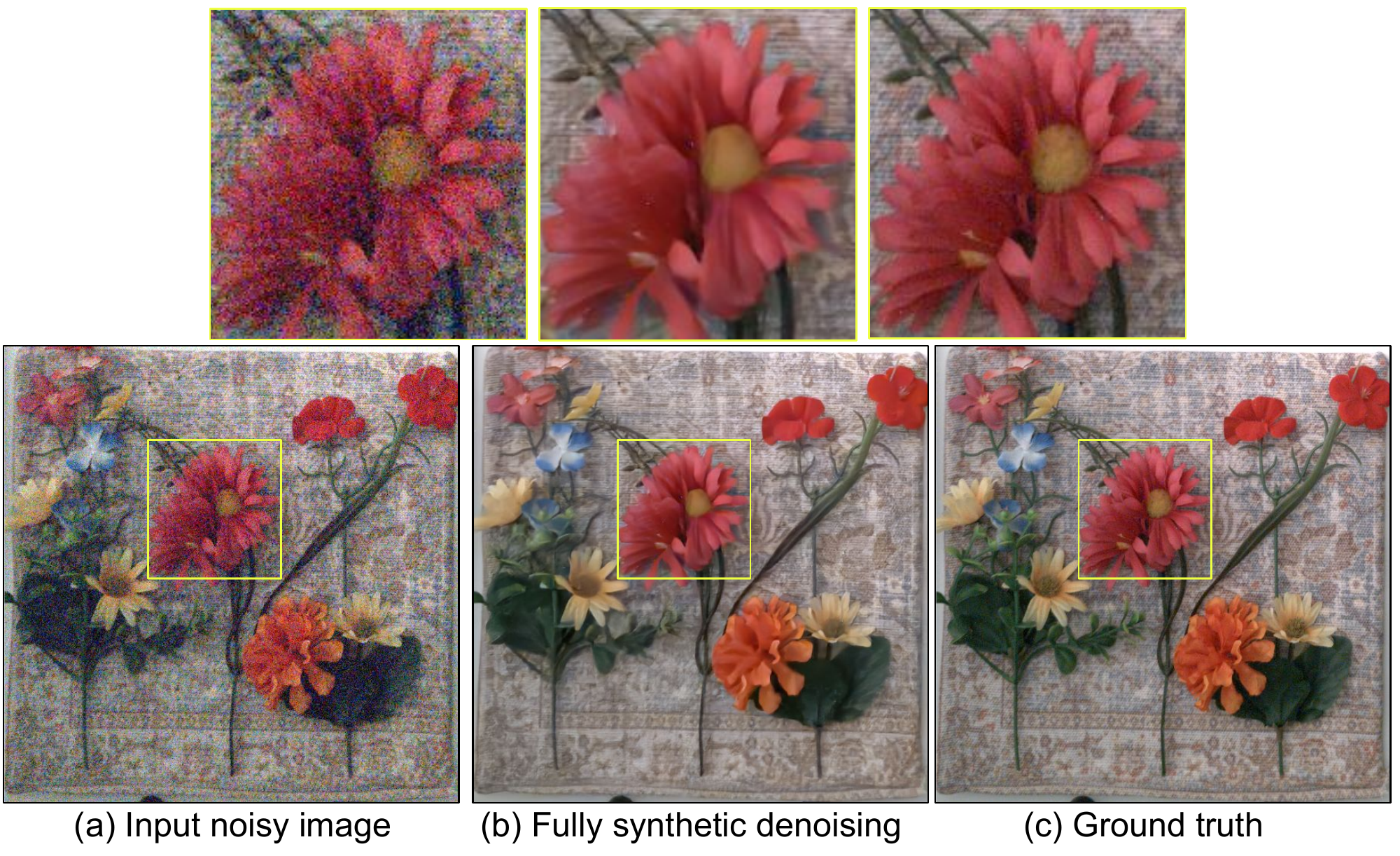}
    \caption{Denoising result visualization. (a) Noisy image captured from Pixel 6 camera under low light (b) Denoised result using our fully synthetic denoising dataset (FSID) (c) Ground truth clean image}
    \label{fig:teaser}
    \vspace{-3mm}
\end{figure}

For this, we have developed an Unreal engine-based procedural synthetic data generation pipeline and a dataset that are tailored to the data-driven low-level computer vision applications.
Our procedural synthetic data pipeline consists of the following modules: 1) Input camera configuration(\eg  camera FOV, resolution), asset, and GT parameters for specific computer vision applications. 2) Procedurally generating random scenes; random 3D objects, materials, geometric transformations, lighting, background, camera motions, object motions, \etc 3) Analysis of the color and textures, noise profile calibration step, and inject the calibrated noise profiles onto the clean ground truth synthetic images. 4) image denoising network training with the clean/noisy synthetic pairs, and validate onto the real captured noisy images. 
In this paper, to demonstrate the effectiveness of our dataset generation pipeline, we train and validate our fully synthetic denoising datasets with a state-of-the-art AI denoiser~\cite{chen2022simple}, and show good denoising results on real noisy images captured from the latest smartphone cameras.
One might ask if any real captured images from the internet can be also used for the clean ground truth images. 
However, compiling many real images from the internet often yields privacy issues (human faces or bodies), scalability, and image quality issues (resolution, noise) \etc. In this paper, we conduct comparisons where our synthetic clean images and real internet images are used in the denoiser training in turn, and we demonstrate that the fully synthetic denoising pipeline performs accurately as real denoising does.
With our fully synthetic dataset, we obtained the benefits of dataset scalability, mitigation of privacy concerns, and avoidance of redundant data of internet-compiled images.

\begin{figure*}[t]
    \includegraphics[width=0.98\textwidth]{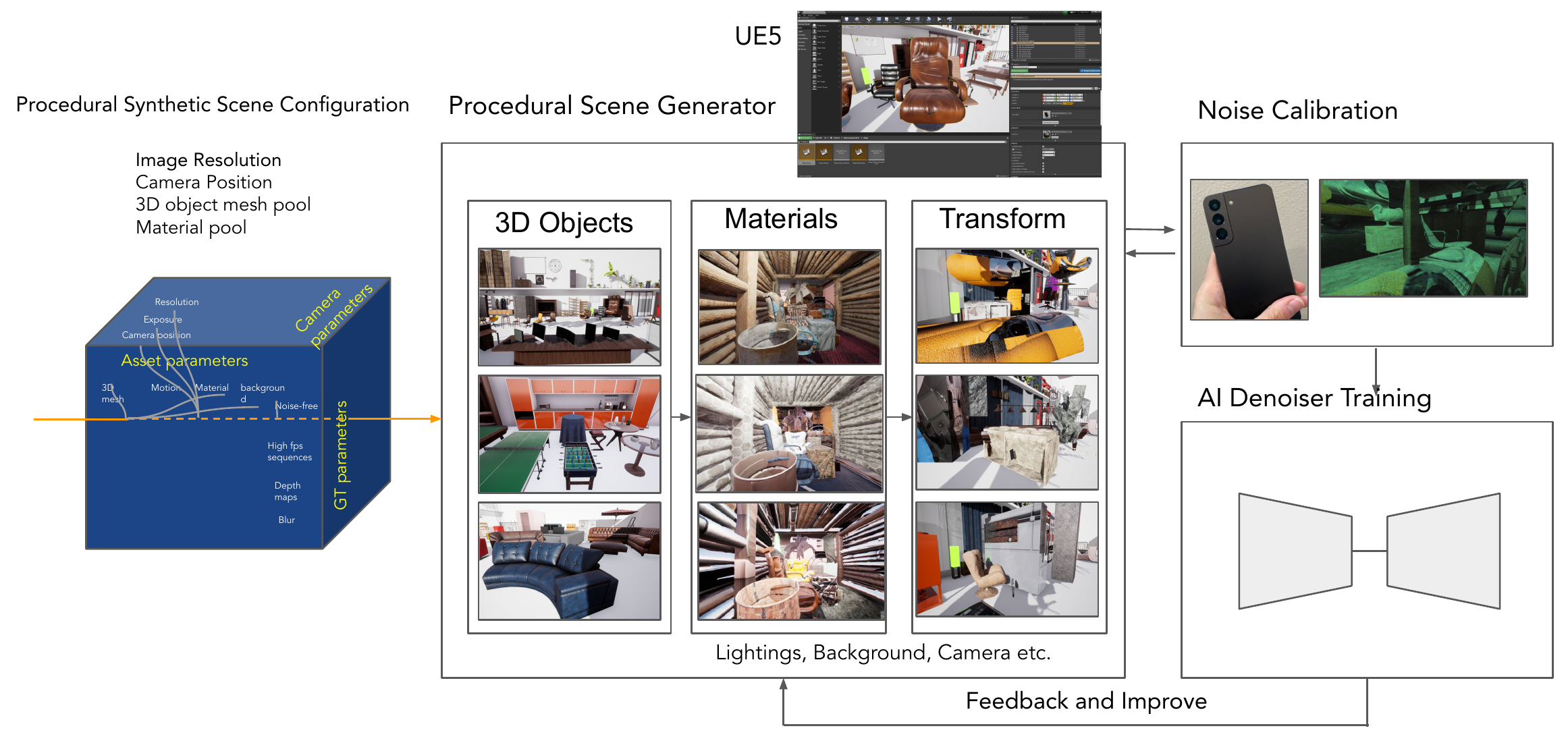}
    \caption{Our procedural synthetic scene generation and synthetic denoising pipeline with Unreal Engine 5 (UE5). Our scene generator combines various 3D objects, materials, and geometric transformations to render diverse textures, colors, and geometries. This random and diverse information plays an important role in low-level vision applications which requires patch-level diversities. Also, our scene generator can control various camera parameters (\eg resolution, FOV, or distortion).}
    \label{fig:psgpipeline}
\end{figure*}

For image denoising, SIDD dataset~\cite{Abdelhamed_2018_CVPR} has been used for benchmarking the various image denoising methods. However, this dataset includes noise profiles of relatively old smartphone devices \eg Galaxy S6 or initial Google Pixel phone. Also because this dataset is a real-captured dataset, the variation of scenes is limited. In our work, we present a fully synthetic image-denoising dataset which provides 175,000  synthetically generated training scenes. Along with the ground truth clean synthetic scenes, we ran a noise calibration method and calibrated the latest smartphone cameras' noise profiles. In this paper, we demonstrate that our fully synthetic image denoising dataset (FSID) is effectively training the SOTA AI denoisers and yields highly accurate denoising performance on the real-captured testing images. 

To summarize, the contributions of our paper are as follows:\\
\begin{itemize}
    \item A procedural synthetic scene generation pipeline that maximizes the randomized color and textures for various low-level computer vision tasks, especially image denoising.
    \item Calibrated noise profiles for latest smartphone cameras \ie Google Pixel 6 and Samsung Galaxy S22. Our fully synthetic pipeline practically allows any newer noise profiles of future cameras.
    \item A fully synthetic image denoising dataset that consists of 175,000 pairs of clean ground truth images and noisy images. We demonstrate that our fully synthetic denoising dataset achieves competitive AI denoising results to that of real image training results.
    \item A test dataset with various materials: \ie color checkerboard, color fabrics/threads, different sizes of texts, a picture, and Money bills, which are captured under four different light settings; 0.5 - 5 lux levels. 
\end{itemize}

\section{Related Works}
\label{sec:relatedworks}
\paragraph{Image Denoising Dataset}
Image denoising is a longstanding problem in computer vision, and there have been many attempts to collect a dataset for training and benchmarking AI-based denoising approaches~\cite{anaya2018renoir,plotz2017benchmarking,Abdelhamed_2018_CVPR,DBLP:journals/corr/abs-1805-01934,wei2020physics,karras2019style,zhang2019poisson,yue2020supervised,ebel2020multisensor,nam2016holistic}.
Conventionally, such a dataset has been synthetically generated by adding sensor-independent white Gaussian noise on existing natural images such as BSD~\cite{martin2001database}.
Recently, some studies collected datasets to evaluate image denoising methods on real images.
RENOIR~\cite{anaya2018renoir} captured pairs of low/high-ISO images to use them as clean and noisy images, respectively.
As the alignment of image pairs is not accurate, DND~\cite{plotz2017benchmarking} proposed a post-processing algorithm including aligning low/high-ISO image pairs to create a high-quality dataset.
SIDD~\cite{Abdelhamed_2018_CVPR} dataset was proposed by capturing 30,000 raw images of 10 distinct scenes with five smartphone cameras and various settings.
It approaches averaging a sequence of images of the same scene to acquire ground truth, but proposed to correct several errors including defective pixels, outlier images, radial distortion, \etc
SID~\cite{DBLP:journals/corr/abs-1805-01934} collected 424 distinct scenes by capturing short/long exposure pairs with two cameras for low-light image denoising.
Even though these datasets have been useful for training and evaluating image-denoising algorithms, they have several limitations such as the limited number of scenes and cameras, redundant data, and privacy concerns when capturing real-world scenarios.
On the other hand, our FSID dataset overcomes the limitations by capturing a dataset using graphics rendering and accurate camera noise calibration.

\paragraph{Synthetic Image for Computer Vision Tasks} 
Due to its usefulness of collecting a large amount of data to fuel AI models, synthetic data is used in many data-driven computer vision applications such as object detection/tracking~\cite{hinterstoisser2018pre,gaidon2016virtual}, pose estimation~\cite{learninghuman,rad2018feature}, semantic segmentation~\cite{tsirikoglou2017procedural,indoorwithnoise,Hu2019SAILVOSSA,8237554}, and optical flow~\cite{butler2012naturalistic,flyingchairs,gaidon2016virtual}.
Even though there exists a discrepancy between real and synthetic data, synthetic data have demonstrated their effectiveness in many tasks.
This is mainly because the data generation process is controllable, thus it is able to synthesize various images including uncommon cases that span the entire space of a task.
In addition, it provides a way to generate data with different levels of complexity~\cite{butler2012naturalistic}, which enables data-driven methods to better learn task-specific knowledge.
There also have been several attempts to narrow the gap between real and synthetic data~\cite{DBLP:journals/corr/SaitoWHNL16,applesynthetic,hu2020sensor,superres,DBLP:journals/corr/abs-1711-06969,8578394,bridgegap,indoorwithnoise} using image-to-image translation~\cite{applesynthetic}.
Our work is in the same line with those synthetic datasets but focuses on realistic noisy and clean image pairs for image-denoising applications.
We exploit a graphics rendering pipeline to generate a variety of images with different objects, materials, lighting, camera/object motion, \etc
On top of it, we synthetically add noise using an accurate noise model calibrated from various smartphone cameras.

\section{Procedural Synthetic Scene Generation}
\label{sec:proceduralsynthetic}
This section describes our procedural synthetic scene generation pipeline. The main goal of the procedural pipeline is to synthetically generate scenes by maximizing the randomization of the color and texture information of pixels and patches in a scene, which potentially represents sufficient variations for the robust training of the data-driven low-level computer vision models. In~\figref{fig:psgpipeline}, our procedural synthetic scene generation pipeline is shown, which combines various 3D objects, materials, and geometric transformations to render diverse textures, colors, and geometries. This random and diverse information plays an important role in low-level vision applications which requires patch-level diversities, \eg image denoising.
\subsection{Scene Randomization}
For our pipeline, we used Unreal Engine 5 (UE5)~\cite{unreal} for the rendering of the procedural scenes. Our synthetic data pipeline consists of the following modules: 1) Input configuration with camera, asset, and GT parameters for image denoising application. Our pipeline provides users to control the camera resolution (in this paper, we used full HD resolution), camera field-of-view (FOV), the number of cameras (mono, stereo, or multiple cameras), camera motion (jittering) mode, and lens distortion to support various applications. For the scene diversity, we used Unreal engine assets, which consist of various indoor objects meshes including table, couch, monitor, chair, desk, plant \etc and variety of materials such as wood, metal, fabric, leather, plastic \etc Those diverse object meshes \(\mathbf{S}_{i}^{t}\)and materials \(\mathbf{M}_{i}^{t}\) were randomly combined and it yields maximized variations of textures and colors. Note that the $i$ is the subset of the total number of object actors present in a scene $N$, and $t$ is the subset of the total number of scenes being generated $M$.  2) Procedurally generating random scenes; random 3D objects, materials, geometric transformations, lightings, background material (\(\mathbf{B}_{j}^{t}\)), camera motions. Note that the $j$ is the subset of the total number of background regions (floor, wall~\etc) present in a scene $K$. For each scene, we used multiple different object actors spawned with random object mesh with random material combinations. For the geometric transformation \(\mathbf{T}_{i}^{t}\), we used 20 different rotation angles for each of the axes, and randomly selected the rotating angle for each of the objects presenting in a scene.
Each of the procedural scene \(\mathbf{I}^{t}\) is generated by randomly combining those properties, which can be described as: 
\begin{equation*}
\label{eqn:scenecomposition}
   \mathbf{I}^{t} = f(\mathbf{S}_{i}^{t},\mathbf{M}_{i}^{t},\mathbf{T}_{i}^{t},\mathbf{B}_{j}^{t})
\end{equation*}
After an initial version of the synthetic scene dataset is rendered, we ran data validation via
3) Analysis of the color and textures, and add real camera-calibrated noise profiles. 4) AI denoiser model training with the clean/noisy synthetic pairs. By checking out the denoising results with each of the procedural scene dataset version, the procedural scene generation pipeline was iterated to obtain better procedural scenes. The ablation study for this will be discussed in the experimental result section.

\begin{figure}[t]
    \includegraphics[width=0.48\textwidth]{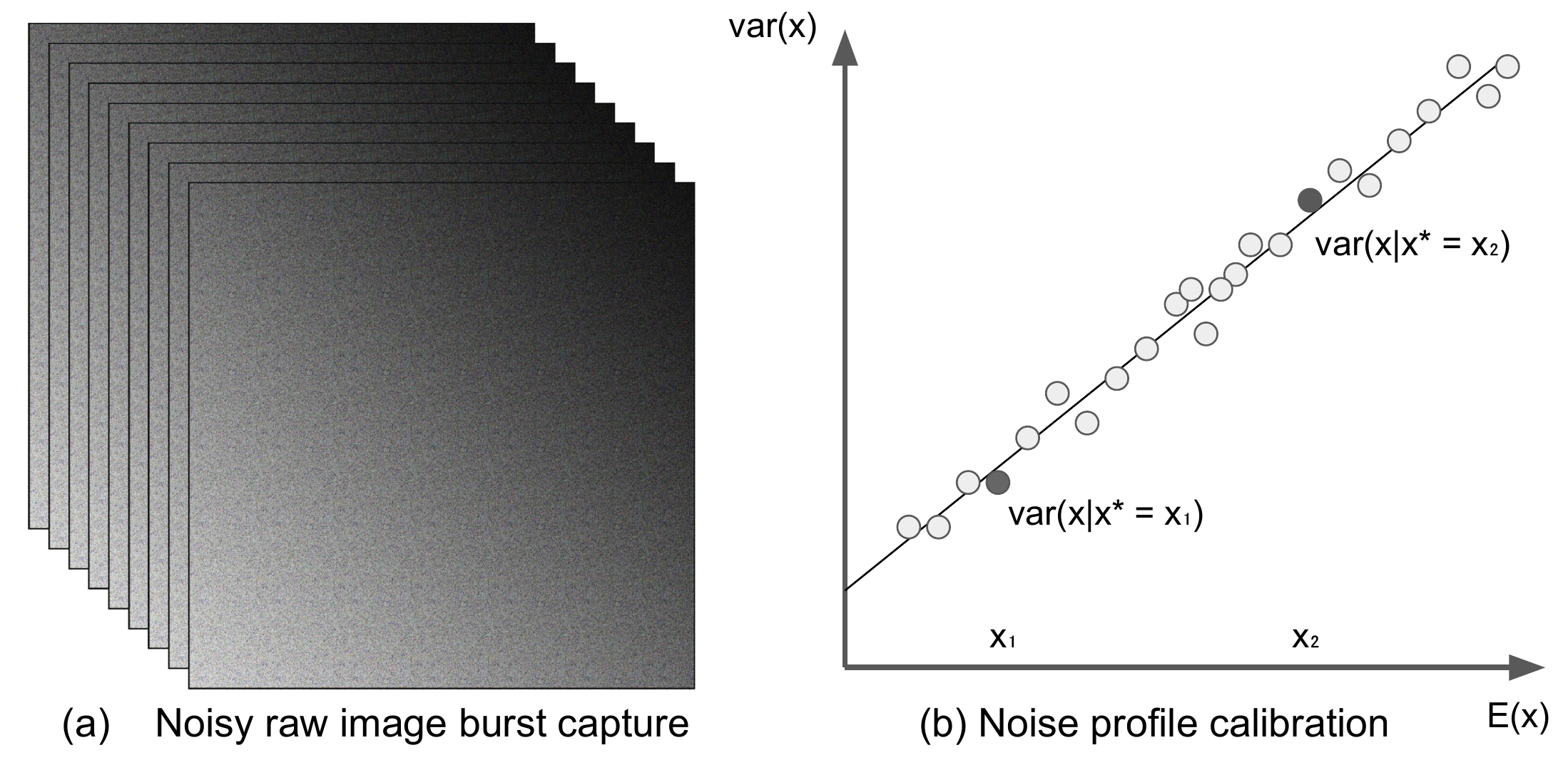}
    \caption{Our noise calibration uses (a) Burst capture of more than 1000 noisy frames. (b) Noise parameter calibration of the linear regression equation~\cite{wang2020}.}
    \label{fig:noisecalibration}
    \vspace{-3mm}
\end{figure}

\begin{figure*}[t]
    \includegraphics[width=0.99\textwidth]{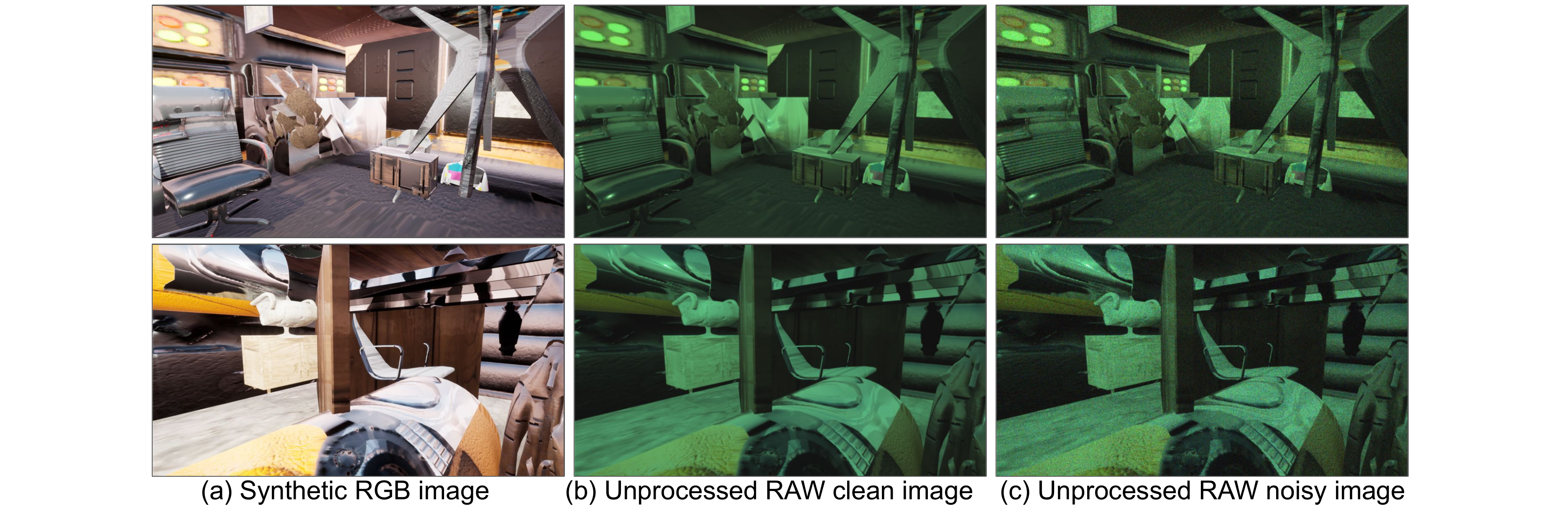}
    \caption{FSID dataset example. (a) Synthetic RGB image generated from Unreal engine (b) RAW clean image from the unprocessing pipeline (c) RAW noisy image with calibrated noise injection. }
    \label{fig:gtnoisyimages}
\end{figure*}

\subsection{Texture and Color Variations}
\label{sec:texturevariation}
The synthetic data generation pipeline is beneficial since it is able not only to generate massive data but also to collect images that resemble real-world scenes without capturing actually capturing them. To represent the real-world scenes from the low-level computer vision perspectives, we verified if our synthetic scenes have enough variations of the textures and colors across the scenes. As an example real image dataset for the baseline, we use the Open Image dataset~\cite{openimages}, and our synthetic dataset is compared against the images in this real dataset. We ran an edge detector with the same threshold setting and computed the ratio of the edges in the image pixels to make sure our synthetic scenes have enough texture variations. 
From the initial version of the synthetic scene dataset, by computing the ratios, the number of the Unreal engine object actors, the combinations of materials, and the combinations of the geometric transformations were set. Also, the background materials \eg floor, ceiling, and wall have been spawned from the background material pool to include enough color and texture information on the background pixels. This is mainly for reducing redundant homogeneous regions in a scene.

\section{Fully Synthetic Denoising Dataset}
\subsection{Noise Profile Calibration}
\label{sec:noisecalibration}
After the synthetic scenes are procedurally generated, the next step is to synthetically model the noisy images by injecting the real noise profile of the camera devices into the clean ground truth images. This section describes the noise calibration method. We follow the basic noise profile calibration method in~\cite{wang2020}. 
Using the smartphone camera device, we capture a series of raw images of a static scene with varying luminance across the scene. The multiple static images from the burst capture were averaged to obtain the mean image $\mathbf{E(x)}$. As presented in the paper~\cite{wang2020}, noise calibration can be expressed into linear regression problem as:
\begin{equation*}
\label{eqn:noise1}
   E(x) = x^* 
\end{equation*}
\begin{equation*}
\label{eqn:noise2}
   Var(x) = kx^* + \sigma^2
\end{equation*}

In~\figref{fig:noisecalibration} (a), the example noisy images for calibrating the noise profile are shown. Our noise calibration step uses the captured raw noisy images. For the specific camera, the noise parameters of the noise equation are estimated. The intensity variations of the noisy frame against the mean images (noise-free clean image) are computed, and those observations are used to get the noise profile of the specific camera device. In~\figref{fig:noisecalibration} (b), The sample points are visualized. From this method, the noise profiles for each smartphone camera are calibrated, and these synthetic noise models are injected into the synthetic clean (denoising ground truth) images.  

\subsection{Synthetic Noisy Image Dataset}
\label{sec:noisydataset}
Using the noise calibration method which was described in the previous subsection, we simulate the synthetic noisy images by injecting the calibrated noise profiles onto the synthetic ground truth clean images. Since our noise model was calibrated by using the Bayer raw images, we unprocess the RGB synthetic images to Bayer raw images, and injected the noise profiles. For the unprocessing pipeline, we followed the method in~\cite{brooks2019unprocessing}. In~\figref{fig:gtnoisyimages} shows some example denoising dataset image pairs. In (a), synthetic RGB images generated from the Unreal engine are visualized. Note that the synthetically rendered scenes from the Unreal engine are completely noise-free, which can be used as ground truth clean images. Most of the real captured images compiled from the internet have various levels of inherent noise, which is challenging to serve as clean ground truth images. In (b), RAW clean images generated from the unprocessing pipeline are visualized. Since our noise modeling and denoising traing is done on Bayer RAW images, these RAW clean images are used as ground truth clean images. In (c), noisy RAW images synthesized by injecting the calibrated noise model are shown. These fully synthetic RAW clean/noisy image pairs are used for training the AI denoiser models. Various noise levels are augmented and used in the training so that it can cover a wide range of noise strengths for each scene. 
In our work, we have modeled noise profiles of the latest smartphone camera devices; Google Pixel 6 and Galaxy S22. 

\section{Image Denoising}
\label{sec:denoising}
Image denoising is a critical component for consumer-level devices, which removes noises inherent to the images or videos captured under dark environments or low-light conditions. Conventional methods use either non-AI-based or data-driven ML model training. For the AI denoisers, previous works have used a handful of real-captured datasets with different exposure and ISO settings to obtain noisy/clean images. Also, previously we could use the real images compiled from the internet \eg Open Image dataset~\cite{openimages}, assume it as a ground truth clean image, and injected noise from the noise model calibrated from camera devices to mimic the noisy images. However, collecting a large-scale dataset of clean images is challenging because of privacy issues, inherent noises, image quality \etc Our pipeline enables us to rely purely on synthetic data generated using our procedural generator, which is used for training and validating the AI denoisers fully synthetically. Note that this dataset generation pipeline can be used not only for image denoising but also for the various low-level computer vision and image processing tasks \ie image super-resolution or video interpolation.

For the AI denoiser training, we generated synthetic clean/noisy pair images. A total of 175,000 image pairs were generated and used for training the denoiser from scratch. Note that our procedural scene generator can automatically generate scenes and there is less limitation for scaling up the number of data.
For the denoiser training, we use our denoising dataset of the latest smartphone cameras. We trained the camera devices' noisy images together with the clean ground truth and demonstrate that the denoiser model effectively performs the denoising of the real-captured noisy images.

\subsection{SOTA Image Denoiser}
\label{sotaimagedenoiser}
To fairly evaluate the influence of different training datasets, we choose the SOTA image denoiser NAFNet~\cite{chen2022simple} as the baseline denoiser\footnote{Source code: \href{https://github.com/megvii-research/NAFNet}{https://github.com/megvii-research/NAFNet}}. NAFNet aims to design a simple baseline for image restoration, by removing nonlinear activation functions or replacing them with multiplication.
It achieves great performance in single image denoising on the SIDD dataset~\cite{Abdelhamed_2018_CVPR}. Besides, it also demonstrates competitive performance on other single-image restoration tasks with relatively low computational complexity.

\begin{table*}
    \small
    \centering
    \begin{tabular}{|c|c|c|c|c|c|c|c|c|}
        \hline
        Lighting setting & \multicolumn{2}{|c|}{0.5 lux} & \multicolumn{2}{|c|}{1 lux} & \multicolumn{2}{|c|}{2 lux} & \multicolumn{2}{|c|}{5 lux} \\\hline
        Procedural Scene Composition & PSNR & SSIM & PSNR & SSIM & PSNR & SSIM & PSNR & SSIM\\\hline\hline
        Random Object & 32.77 & 0.7940 & 32.64 & 0.8351 & 32.24 & 0.8569 & 31.63 & 0.8869 \\\hline
        Random Material & 32.78	& 0.7802 & 32.60 & 0.8306 & 32.01 & 0.8550 & 31.68 & 0.8890\\\hline
        Object + Material & \bftab 32.99 & 0.7920 & 32.74 & 0.8383 & \bftab 33.01 & 0.8623 & 31.69 & 0.8963 \\\hline
        Object + Material + Transformation & 32.97 & \bftab 0.7928 & \bftab 32.76 & \bftab 0.8384 & 32.97 & \bftab 0.8626 & \bftab 31.80 & \bftab 0.8973 \\\hline
    \end{tabular}
    \caption{Ablation study: Our procedural synthetic scenes generation pipeline consists of multiple steps and each step are validated in this table. For each version, 100k image pairs were used for training.}
    \label{tab:ablation}
    \vspace{-2mm}
\end{table*}

\subsection{Conventional Benchmark Dataset}
\label{conventionaldataset}
For the training and validating various AI denoising methods, there has been several benchmark denoising dataset presented. One of the representative datasets for image denoising is Smartphone Image Denoising Dataset (SIDD)~\cite{Abdelhamed_2018_CVPR}, which proposes a pipeline for estimating ground truth from real noisy images. SIDD is composed of ~30,000 noisy images taken from 10 scenes under different lighting conditions. The corresponding ground truth images are estimated by a series of operations including outliers removal and dense alignment to generate the robust mean image. SIDD uses five smartphone cameras: Apple iPhone 7, Google Pixel, Samsung Galaxy S6 Edge, Motorola Nexus 6, and LG G4, which are relatively old. So there has been a need for the latest smartphone cameras' denoising dataset. Our work provides calibrated noise profiles and clean/noisy datasets for Google Pixel 6 and Samsung Galaxy S22. We provide the fully synthetic 175,000 denoising dataset where each scene has various contents. Our pipeline also allows newer camera devices' noise profiles to be practically calibrated and updated, which could be a scalable solution for future extensions. Since our dataset models the newer smartphone devices, we do not directly compare the denoising performance against the SIDD~\cite{Abdelhamed_2018_CVPR}. To do the fair comparison, we construct a comparison dataset by injecting the Pixel 6 and Galaxy S22 noise profile onto the conventional real dataset~\cite{openimages}. Our fully synthetic denoising dataset is compared to this dataset.

\section{Experimental Results}
\label{experiments}

For evaluating the effectiveness of the fully synthetic denoising pipeline and the dataset, we conducted experiments for validation and testing. Firstly we have captured a testing dataset with the smartphone devices. Second, we conducted an ablation study to validate the procedural scene composition. The various versions of the training dataset with different scene compositions (the same amount of training data for each version) have been trained with the NAFNet denoiser. After the training is done, the trained AI denoiser models are used to test the denoising of the real-captured testing dataset. The following subsection will describe more details on it. Third, to demonstrate the effectiveness of the fully synthetic denoising pipeline, we also compared the fully synthetic denoising performance with real image denoising performance. For the real image dataset, Open Image~\cite{openimages} dataset was used as the training dataset. The synthetic dataset and real dataset were trained separately (same amount of training data, 175k pairs) and tested with the same testing dataset. The comparison of the result denoising images between the fully synthetic training model and the real training model is shown both qualitatively and quantitatively.

\subsection{Test dataset}
For validating our synthetic training models and comparing denoising results against the conventional dataset, we used the latest smartphone cameras- Google Pixel 6 and Samsung Galaxy S22 to capture the testing dataset. 
For the capturing light setup, We set the ambient light setting with 4 different lux levels; 0.5, 1.0, 2.0, and 5.0 lux. The denoising accuracy has been verified across these various lighting settings.  
For the contents of the testing scenes, 10 different objects/materials were used- calibration star chart, bar chart, color chart, text written in various different font sizes, money bills and coins, color patches in fabrics, color threads, and a picture (magazine), plastic flower and plant leaves. We have verified the denoising accuracy with each of the materials.  

\begin{figure}[t]
    \includegraphics[width=0.49\textwidth]{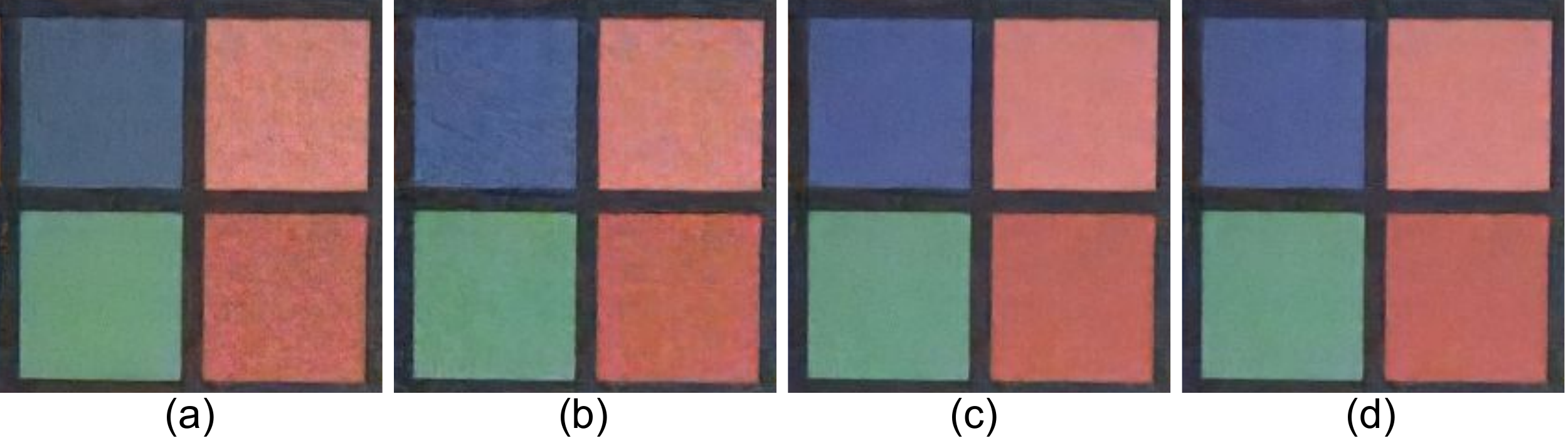}
    \caption{Ablation study: Denoising result comparison with different scene compositions. (a) Randomized object, (b) Randomized material, (c) Combination of randomized object and material, (d) a Combination of the randomized objects, materials and geometric transformation.}
    \label{fig:ablationchecker}
    \vspace{-3mm}
\end{figure}

\begin{figure*}[t]
    \includegraphics[width=0.99\textwidth]{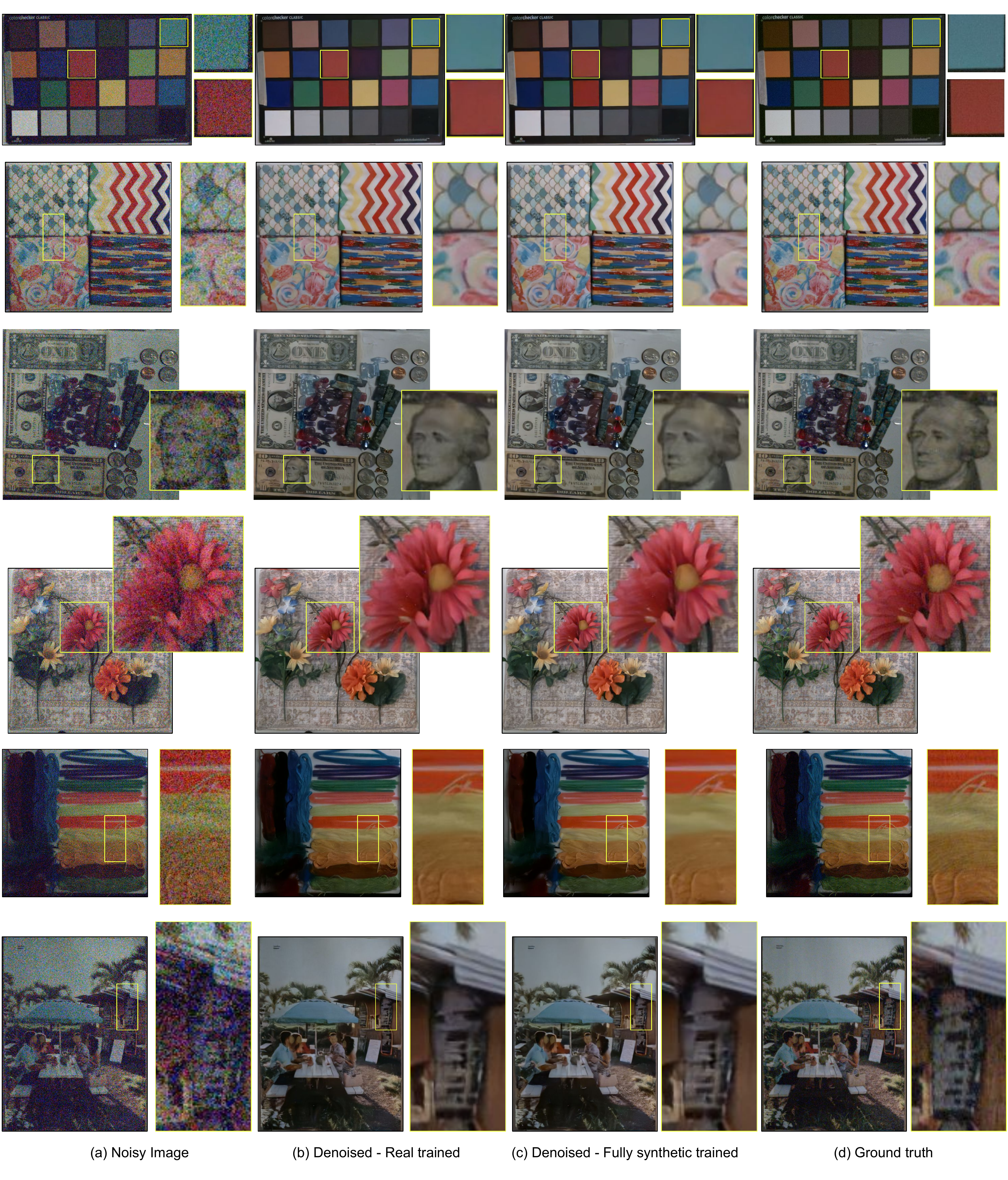}
    \caption{Image denoising results comparison- From the left, (a) captured noisy image, (b) denoised image using Open image dataset trained model, (c) denoised image using our FSID dataset trained model and (d) ground truth}
    \label{fig:denoisingresultallscenes}
\end{figure*}

\begin{figure*}[t]
    \includegraphics[width=0.99\textwidth]{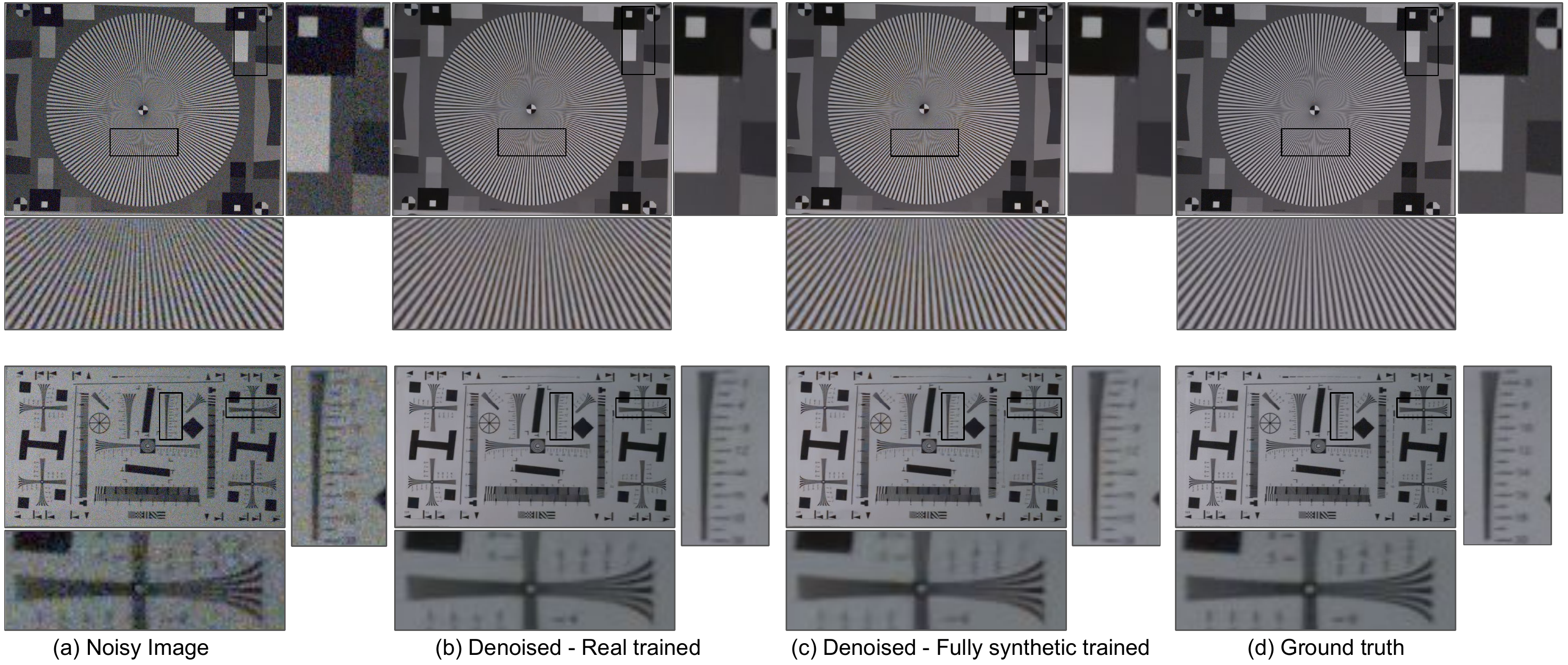}
    \caption{Image denoising results comparison- Using synthetic training model and real training model.}
    \label{fig:denoisingresultcharts}
    \vspace{-1mm}
\end{figure*}

\begin{table*}
    \small
    \centering
    \begin{tabular}{|c|c|c|c|c|c|c|c|c|}
        \hline
        Lighting setting & \multicolumn{2}{|c|}{0.5 lux} & \multicolumn{2}{|c|}{1 lux} & \multicolumn{2}{|c|}{2 lux} & \multicolumn{2}{|c|}{5 lux} \\\hline
        Denoiser Training Dataset & PSNR & SSIM & PSNR & SSIM & PSNR & SSIM & PSNR & SSIM\\\hline\hline
        Real dataset (Open Image) & 31.44 & 0.7430 & 32.66 & 0.8243 & 33.11 & 0.8800 & 30.69 & 0.8095 \\\hline
        Synthetic dataset (FSID) &  31.04 & 0.7376 &  31.92 & 0.8213 & 33.12 & 0.8788 & 30.55 & 0.8100\\\hline
    \end{tabular}
    \vspace{-1mm}
    \caption{Comparison of the fully synthetic image-trained denoising results against real image-trained denoising results: Pixel6.}
    \label{tab:comp_pixel6}
\end{table*}

\subsection{Ablation Study}
To validate each of the steps of our procedural synthetic scene generation pipeline, an ablation study was conducted. We have generated 4 different versions (v1-v4) of the synthetic scene composition method and compiled the training dataset: (1) Randomized object, (2) Randomized material, (3) Combination of randomized object and material, (4) Combination of randomized objects, materials and geometric transformation, The each dataset version includes the same amount of data - 100k synthetic image pairs of clean/noisy and same image resolution for the fair comparison. Each version of the training dataset with different scene compositions was trained with the NAFNet. After the multiple trainings were done, the trained AI denoiser models were used to test the denoising of the real-captured testing dataset. The result comparison is visualized in~\figref{fig:ablationchecker}. The denoising accuracy is higher in (c) and (d) compared to those in (a) and (b). Also, the results images are evaluated with PSNR and SSIM measures as seen in~\tabref{tab:ablation}. The PSNR/SSIM were measured for each lux. As seen in the table, we can see the PSNR and SSIM scores are higher when both the random object and random materials were combined than solely controlling the object asset or material. The geometric transformation also increases the denoising performance overall through the different lux levels.

\subsection{Comparison against real-trained model}
To verify the accuracy of the fully synthetic denoising pipeline, we also compared the fully synthetic denoising results against real image denoising results. To validate that the fully synthetically generated procedural scenes provide sufficient texture and color information for the image denoising, a real image dataset \ie Open Image~\cite{openimages} dataset was used as a compared training dataset. The 175k pairs of the synthetic dataset and real dataset were trained separately and tested with the same testing dataset respectively. The denoising results of the two models are compared and evaluated both qualitatively and quantitatively.
In~\figref{fig:denoisingresultallscenes} and~\figref{fig:denoisingresultcharts}, from the left, original noisy images captured from the Pixel 6 device, denoised results using real image training model, denoised results using our synthetic training model and ground truth clean images are visualized. The test images show that our fully synthetic trained model yield visually identical denoising quality compared to the real-trained model. 

Also, we conducted a quantitative evaluation as in~\tabref{tab:comp_pixel6}. The PSNR and SSIM scores are measured for each test scene under different lux levels respectively. As seen in the table, throughout the different lux levels, we can observe that our synthetic denoised results perform closely to the real denoised results. From the less noise of 5 lux- PSNR: 30.69 vs 30.55, SSIM: 0.8095 vs 0.8100) to the highest noise of 0.5 lux- PSNR: 31.44 vs 31.04, SSIM: 0.7430 vs 0.7376), our fully synthetic denoising results achieve comparable scores with those of the real denoising results.

\section{Conclusion}
In this work, we have demonstrated that our proposed fully synthetic denoising method achieves good denoising quality both in qualitative and quantitative analysis. Our synthetic denoising dataset generation pipeline included randomized color, texture, and geometry combinations to ensure diversity of the scenes, which is conducive to training various low-level computer vision ML models \eg image denoiser. By procedurally generating a large number of scenes, calibrating the real noise profiles, and using them to train the CNN denoiser, we could achieve accurate denoising results which yield almost identical quality to real image-trained results.
For future work, our pipeline can be further extended to various other low-level computer vision tasks \eg image super-resolution or deblurring.

{\small
\bibliographystyle{ieee_fullname}
\bibliography{egbib}
}

\end{document}